\documentclass[10pt, a4paper]{article}
\usepackage{lrec}
\usepackage{graphicx}
\usepackage{tabularx}
\usepackage{soul}

\usepackage{epstopdf}
\usepackage[latin1]{inputenc}

\usepackage{hyperref}
\usepackage{xstring}

\usepackage{times}
\usepackage{latexsym}
\usepackage{multirow}
\usepackage{enumitem}
\usepackage{amssymb}
\usepackage{graphicx}

\newcommand{\ie}{\textit{i.e.}}
\usepackage{ascmac}

\usepackage{url}
\usepackage{color}
\usepackage{colortbl}
\usepackage{subfigure}

\title{Constructing a Visual Relationship Authenticity Dataset}

\name{Chenhui Chu, Yuto Takebayashi, Mishra Vipul, Yuta Nakashima}

\address{Osaka University \\
        Yamada-oka 2-8, Suita, Osaka, Japan \\
         \{chu, n-yuta\}@ids.osaka-u.ac.jp, \{mishra.vipul, takebayashi.yuto\}@ist.osaka-u.ac.jp\\
         }

\abstract{
A visual relationship denotes a relationship between two objects in an image, which can be represented as a triplet of (subject; predicate; object). Visual relationship detection is crucial for scene understanding in images. Existing visual relationship detection datasets only contain {\it true} relationships that correctly describe the content in an image. However, distinguishing {\it false} visual relationships from true ones is also crucial for image understanding and grounded natural language processing. In this paper, we construct a visual relationship authenticity dataset, where both true and false relationships among all objects appeared in the captions in the Flickr30k entities image caption dataset are annotated. The dataset is available at \url{https://github.com/codecreator2053/VR_ClassifiedDataset}. We hope that this dataset can promote the study on both vision and language understanding. 
\\ \newline  }

\begin{document}

\maketitleabstract

\section{Introduction}
\begin{figure*}[t]
  \begin{center}
     \includegraphics[width=0.7\hsize]{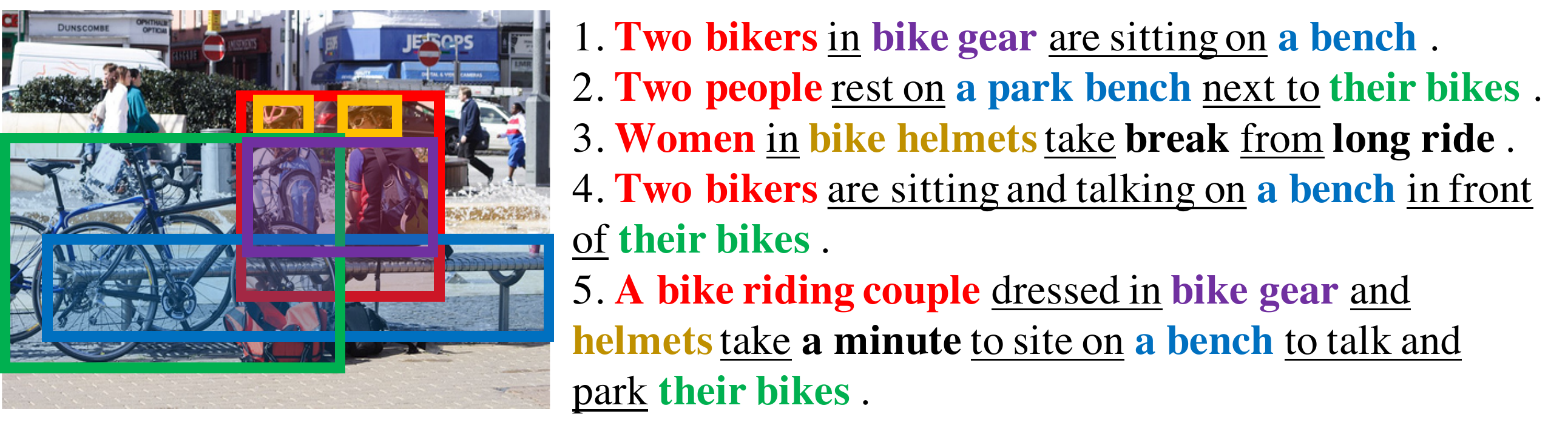}
     \caption{An example from the Flickr30k dataset. There are 5 captions for each image. Objects/entities in each caption are localized to corresponding image regions in the image (shown in the same color). The underlined phrases denote the predicates among the objects. There can be both true and false visual relationships among the objects. For example, ``two bikers in bike gear'' is a true visual relationship, while ``bike gear are sitting on a bench'' is a false one.}
     \label{fig:flickr30k}
  \end{center}
\end{figure*}

A visual relationship is any relationship describing the interaction between two objects in an image \cite{DBLP:conf/cvpr/SadeghiF11,7298990,lu2016visual,krishnavisualgenome}.
There are various visual relationships between the objects in an image. For example, in Figure \ref{fig:flickr30k}, the relationship between ``two bikers'' and ``a  bench'' is ``are sitting on'' or ``are sitting and talking on.'' Similarly, the relationship between ``two bikers'' and ``bike gear'' is ``in'' or ``dressed in.'' Detecting such visual relationships can make the understanding of the entire image possible. 

Four datasets have been published for visual relationship detection between objects in an image, i.e., Visual Phrases \cite{DBLP:conf/cvpr/SadeghiF11}, Scene Graph \cite{7298990}, Visual Relationship Detection (VRD) \cite{lu2016visual},\footnote{{https://cs.stanford.edu/people/ranjaykrishna/vrd/}} and Visual Genome (VG) \cite{krishnavisualgenome}.\footnote{{https://visualgenome.org/}} In these existing datasets, visual relationships are represented as triplets of (\textit{subject}; \textit{predicate}; \textit{object}). \textit{Subject} and \textit{object} are two different objects in an image, and \textit{predicate} can be either a preposition that describes the positional relationship such ``under" or ``in front of," or verbal relationship such as ``hold" or ``ride."

A problem with the Visual Phrases dataset is that the number of types of visual relationships is limited (i.e., only 13 relationship types). The Scene Graph, VRD, and VG datasets have adequate numbers of types of visual relationships, but the annotations are limited to {\it true} visual relationships. We define a visual relationship as {\it true} if it correctly describes the content in an image; otherwise the relationship is {\it false}. For example, in Figure \ref{fig:flickr30k}, ``two bikers in bike gear'' is a true visual relationship, while ``bike gear are sitting on a bench'' is a false one.
However, distinguishing {\it false} visual relationships from true ones is also crucial for image understanding as well as for some grounded natural language processing (NLP) tasks. For example, in visually grounded paraphrase identification \cite{Chu_COLING18}, false visual relationships should be excluded from valid paraphrase candidates. Moreover, the correct detection of true and false visual relationships between objects represented by noun phrases, can improve dependency parsing of the noun phrases.

In this paper, we construct an authenticity dataset for visual relationships. We annotate the true and false visual relationships among all objects/entities appeared in the captions of the Flickr30k entities image caption dataset \cite{Plummer_2015_ICCV}. Our annotation has been done in two stages: Firstly, we list all visual relationships to be annotated in an image and use dependency parsing to detect the true relationships; Secondly, we annotate the relationships that could not be detected by dependency parsing via crowdsourcing.
As a result, we successfully construct a visual relationship dataset with both true and false relationships, which covers relationships among all objects that appeared in the captions in the Flickr30k dataset. This dataset can not only promote the study on image understanding but also contribute to NLP. 

\begin{table}[t]
\small
    \centering
    \begin{tabular}{l|r|r|r|r} \hline
     & Images & Rel. Types & Rel. Ins. & Auth.\\ \hline
     Visual Phrases & 2,769 & 13 & 2,040 & No \\ 
     Scene Graph & 5,000 & 23,190 & 109,535 & No \\ 
     VCD & 5,000 & 6,672 & 37,993 & No  \\ 
     VG & 108,077 & 42,374 & - & No \\ \hline
    Ours & 31,769 & 24,819 & 266,423 & Yes  \\ \hline
    \end{tabular}
    \caption{\label{tab:statistic_comp} Statistical comparison of our visual relationship dataset against previous datasets. ``Auth.'' denotes authenticity,  ``Rel.'' denotes relationships, and ``Ins'' denotes instances.}
\end{table}

\section{Related Work}
\subsection{Visual Relationship Datasets}
The Visual Phrases dataset \cite{DBLP:conf/cvpr/SadeghiF11} contains 17 phrases and 8 objects on the Pascal VOC2008 dataset \cite{Everingham:2010:PVO:1747084.1747104}. The Scene Graph dataset consists of not only visual relationships but also the attributes of objects \cite{7298990}. Their dataset is mainly constructed for image retrieval, which contains 109,535 instances of relationships in 5,000 images. The VRD dataset contains 100 object categories and 70 predicates for 5,000 images \cite{lu2016visual}.
VG is a large-scale dataset containing more than 100k images, annotated with various labels such as objects, attributes, relations, and scene graphs \cite{krishnavisualgenome}. 
Being different from these existing datasets, our dataset is the only one that annotates the authenticity of visual relationships. A statistical comparison of our dataset and previous ones are shown in Table \ref{tab:statistic_comp}.

\subsection{Visual Relationship Detection Models}
The most challenging problem in visual relationship detection is the data sparseness of relationships, and many studies have been conducted to address this problem.
\newcite{lu2016visual} proposed a model that handles objects and predicates independently. They then combined them for visual relationship detection. \newcite{Yu_2017_ICCV} used linguistic knowledge from both the dataset and Wikipedia to regularize
the visual model. \newcite{Plummer_2017_ICCV} used various visual features, such as the appearance, size, position, attribute of objects, and spatial relationships between objects. \newcite{Zhuang_2017_ICCV} treated the subject and object in a visual relationship as context, and the predicate as interaction. They modeled the interaction with context in multiple relationships for visual relationship detection.
\newcite{DBLP:journals/corr/abs-1811-00662} pointed out the importance of language bias and spatial features, and fused them for visual relationship detection. \newcite{Peyre_2017_ICCV}
proposed a weakly-supervised model learnt from image-level labels. \newcite{Zhang_2017_ICCV} simultaneously selected object pairs and classified them for both object and relationship detection. \newcite{Hwang_2018_CVPR} 
first learnt a prior by a multi-relational learning model, and then used a factorization scheme for the prior.
\newcite{Liao_2019_CVPR_Workshops} used a recurrent neural network to model the semantic connection among objects. \newcite{Peyre_2019_ICCV} transferred visual phrase embeddings from triplets in the training data to unseen test triplets using analogies between relationships containing similar objects. \newcite{Zhan_2019_CVPR} used unlabeled relationships to improve the accuracy of visual relationship detection.


\section{Dataset Construction}
\begin{figure}
    \centering
    \includegraphics[width=\hsize]{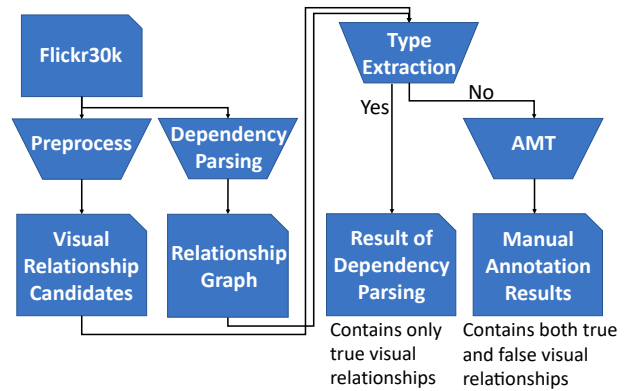}
    \caption{The pipeline for dataset construction. Our dataset consists of the results from dependency parsing and crowdsourcing.
    }
    \label{fig:flowchart}
\end{figure}

The pipeline for constructing our visual relationship authenticity dataset is shown in Figure \ref{fig:flowchart}. Firstly, we preprocess the captions from the Flickr30k dataset to generate all possible visual relationship candidates. Next, we apply dependency parsing to construct a dependency relationship graph among the entities and the predicates. After that we propose to apply {\it type extraction} to detect true relationships among the candidates. The detected ones must belong to both the predefined types in the relationship graph and the visual relationship candidates. Furthermore, the visual relationship candidates that are not detected by type extraction are further annotated via crowdsourcing.

\subsection{The Flickr30k Entities Dataset}

Our dataset is based on the Flicker30k entities dataset \cite{Plummer_2015_ICCV}. Flickr30 entities is a large-scale dataset with images and captions, in which the entities in the captions are localized to the regions in the images. Following \cite{Plummer_2015_ICCV}, we use 29,769, 1,000, and 1,000 samples as training, validation, and testing splits, respectively.
An example of an image and its captions from the Flickr30k entities dataset is shown in Figure \ref{fig:flickr30k}.
Each image has 5 captions, and the entities in the captions are localized by their corresponding regions in the image. In this example, all entities in the captions have corresponding regions in the image; however, there can be more abstract entities related to events and scenes, such as ``break'' and ``a minute'' that do not have corresponding visual concepts and thus have no corresponding regions in the image.

\subsection{Visual Relationship Candidate Generation}
\label{sec:preprocess}
In the Flickr30k entities dataset, entities in captions are annotated by chunking noun phrases in the Flickr30k captions \cite{Young:TACL:2014}. These entities are further categorized into types such as people, body parts, and clothing etc. using a manually constructed dictionary \cite{Plummer_2015_ICCV}. As shown in Figure \ref{fig:candidate}, visual relationship candidates are generated by combining two entities in a caption. Specifically, all pairs of entities in a single caption are combined as candidate relationships, where the subjects and objects must be the ones in which they appear in the caption. 
We define the phrase lying between two entities as predicate. More specifically, the phrase that comes right before the second entity in the caption is treated as the predicate.
In Figure \ref{fig:candidate}, three visual relationship candidates are generated.

\begin{figure}
\begin{screen}
$[$/EN$\#1$/people Two bikers$]$ in $[$/EN$\#2$/other bike gear$]$ are sitting on $[$/EN$\#3$/other a bench$]$.\\
\centering $\Downarrow$
\begin{enumerate}
\setlength{\parskip}{0cm} 
\setlength{\itemsep}{0cm} 
    \item Two bikers in bike gear
    \item Two bikers are sitting on a bench
    \item bike gear are sitting on a bench
\end{enumerate}
\end{screen}
\vspace{-6mm}
\caption{Visual relationship candidate generation.}
\vspace{+2mm}
\label{fig:candidate}
\end{figure}

Some phrases lying between two entities like ``,'', ``, and'', ``while'' and ``space'' are not predicates and thus are removed from the group of candidates.
Entities in the dataset have type tags describing their characteristics (such as ``people'' or ``clothing''). The entities having the type tag ``notvisual'' do not qualify as candidates for visual relationships, because they are not visual in the image. The statistics of the generated visual relationship candidates are shown in the first row of Table \ref{tab:statistic}.


\begin{table}[t]
    \centering
    \begin{tabular}{l|l|r|r|r} \hline
     & Auth. & Train & Validation & Test\\ \hline
    \# Candidates & N/A & 249,706 & 8,425 & 8,292 \\\hline
    \# DP Rel. & True & 90,665 & 3,015 & 2,933 \\ \hline
    \# Rel. & True & 91,608 & 3,455 & 3,299 \\ 
    \ \ \ by AMT & False & 67,433 & 1,955 & 2,060 \\ \hline
    \end{tabular}
    \vspace{-1mm}
    \caption{\label{tab:statistic} Numbers of visual relationships obtained in different steps of our dataset construction pipeline. ``Auth.'' denotes authenticity,  ``Rel.'' denotes relationships, and ``DP'' denotes dependency parsing.}
    \vspace{-1mm}
\end{table}

\subsection{Relationship Detection via Dependency Parsing}
\label{sec:parser}
If a caption correctly represents image content and has correct dependency relations, we can detect true visual relationships based on the dependency relationship between two entities. In this section, we describe the method for detecting true visual relationships via dependency parsing. 
Dependency parsing of the captions is done by the Stanford parser.\footnote{https://nlp.stanford.edu/software/lex-parser.shtml} The parsed captions are cross checked with the candidates for visual relationships obtained in Section \ref{sec:preprocess} and then true visual relationships are extracted. 

In detail, the detection of true visual relationships are carried out in the following manner.
\begin{enumerate}
\setlength{\parskip}{0cm} 
\setlength{\itemsep}{0cm} 
    \item Do dependency parsing on the captions and construct a directional graph, of which nodes are words.\footnote{Collapsed-dependencies are treated as a directional graph.}
    \item Merge nodes in the graph based on the entities and predicates in Flickr30k. The edges connecting the same nodes are ignored.
    \item Get the visual relationship if the entities and their predicate belong to predefined types. 
    \item Extract the visual relationships obtained in Step 3 as true visual relationship if they are also contained in the candidates generated in Section \ref{sec:preprocess}
\end{enumerate}

Steps 3 and 4 are named as type extraction. The ``predefined types'' in Step 3 refer to the dependency relationship of the entity pair and their predicate. As dependency parsing is not error-free, we empirically decide the combination of types. Let EN1 (\ie, subject) and EN2 (\ie, object) be the pair of entities and RE be the predicate between them. As long as EN1, EN2, and RE are linked in the graph, the combinations that make a valid visual relationship are the following 7 types. 
\begin{itemize}
\setlength{\parskip}{0cm} 
\setlength{\itemsep}{0cm} 
    \item A: RE $\rightarrow$ EN1, RE $\rightarrow$ EN2
    \item B: EN1 $\rightarrow$ RE, EN1 $\rightarrow$ EN2
    \item C: EN1 $\rightarrow$ PRE, EN2 $\rightarrow$ RE
    \item D: RE $\rightarrow$ EN1, EN2 $\rightarrow$ EN1
    \item E: EN1 $\rightarrow$ RE $\rightarrow$ EN2
    \item F: RE $\rightarrow$ EN1 $\rightarrow$ EN2
    \item G: EN1 $\rightarrow$ EN2 $\rightarrow$ RE
\end{itemize}
A and B are cases of having a common parent node, and C and D are cases of having a common child node.

\begin{figure}[t]
    \centering
    \includegraphics[viewport=0 0 668 297, width=8.0cm]{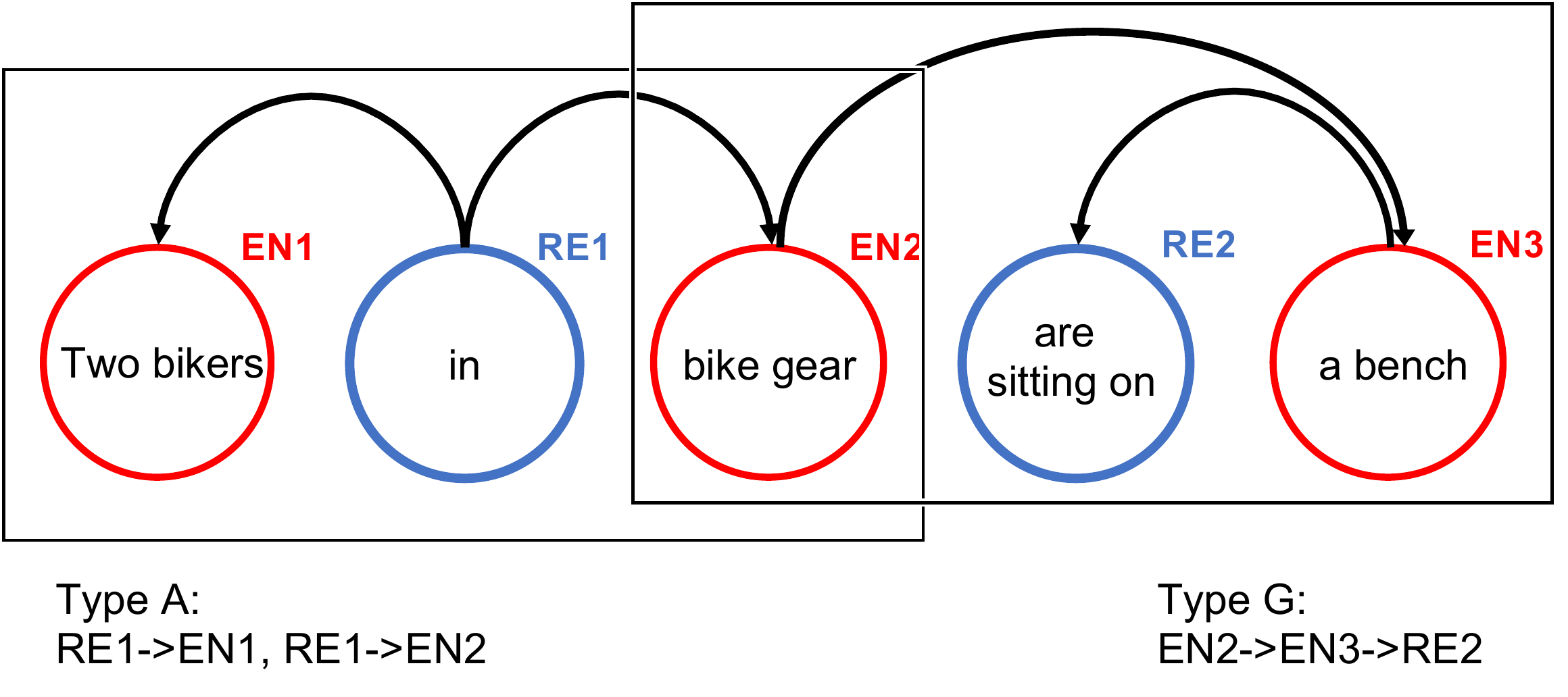}
    \caption{A graph with nodes merged according to entities and predicates. 
     Note that the visual relationship of ``Two bikers are sitting on a bench'' cannot be extracted in this graph.}
    \label{fig:parse_graph}
\end{figure}

Figure \ref{fig:parse_graph} shows an example of a graph built by the dependency parsing with entities and predicates being merged into nodes for the caption ``$[$/EN$\#1$/people Two bikers$]$ in $[$/EN$\#2$/other bike gear$]$ are sitting on $[$/EN$\#3$/other a bench$]$." In this example, ``Two bikers in bike gear'' and ``bike gear are sitting on a bench'' are extracted as types A and G, respectively. Note that the latter candidate is an erroneous extraction due to parsing errors. Likewise, ``two bikers are sitting on a bench'' is a valid candidate but cannot be extracted because it does not match any of the types.

To see which types contribute to extraction, we annotated 186 visual relationship candidates from the validation data by ourselves for the evaluation. The precision/recall/F-score when either one of A to G or some combinations of them are used is shown in Table \ref{tab:Pre/Rec}. Note that the B, D, and F types yielded neither true positives nor true negatives, so their scores could not be calculated. 
The combinations were decided upon by combining the most precise types. Because of the high precision and recall of 90.5\% and 49.7\% respectively, we use A, E and G for combination of types. 
The visual relationships within the 186 manually annotated data that can be extracted as type C were also included in type E. 
The number of visual relationships obtained by dependency parsing are presented in the second row of Table \ref{tab:statistic}.
\begin{table}[t]
    \centering
    \begin{tabular}{@{}l|r r r|r r r@{}} \hline
Type & TP & FP & FN & Precision & Recall & F-score \\ \hline
A & 15 & 1 & 138 & 93.8 & 9.8 & 17.8 \\
C & 5 & 1 & 148 & 83.3 & 3.3 & 6.3 \\
E & 41 & 7 & 112 & 85.4 & 26.8 & 40.8 \\
G & 23 & 1 & 130 & \bf{95.8} & 15.0 & 26.0 \\ \hline
AG & 37 & 2 & 116 & 94.9 & 24.2 & 38.5 \\
AEG  & 76 & 8 & 77 & 90.5 & \bf{49.7} & \bf{64.1} \\
ACEG & 76 & 8 & 77 & 90.5 & \bf{49.7} & \bf{64.1} \\ \hline
    \end{tabular}
    \caption{\label{tab:Pre/Rec}Precision/recall/f-score when either one of types A-G or their combinations are used.
    }
\end{table}


\subsection{Annotation with AMT}
\begin{figure}[t]
    \centering
    \includegraphics[width=\hsize]{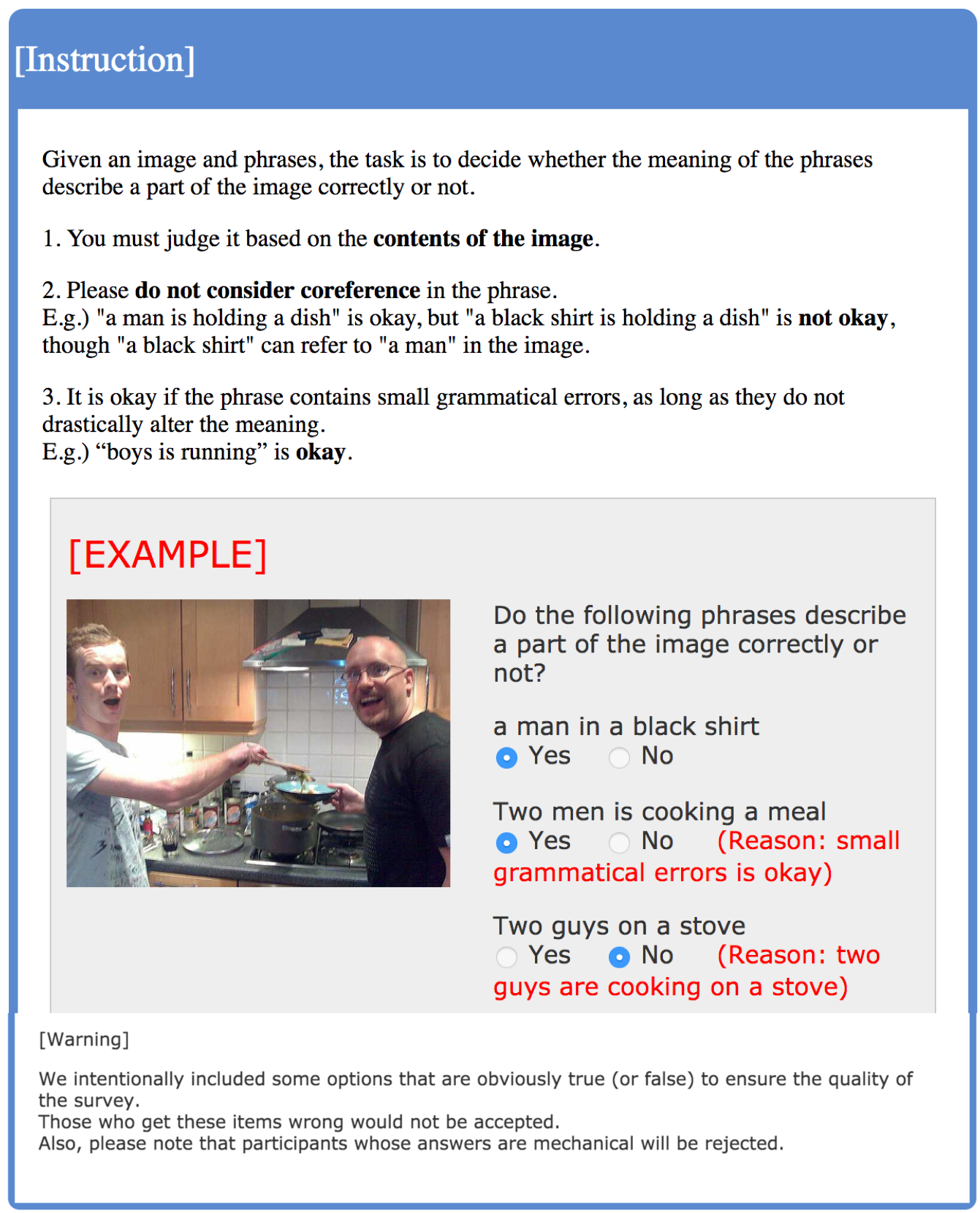}
    \vspace{-5mm}
    \caption{The instruction part in our interface.}
    \vspace{3mm}
    \label{fig:instruction}
\end{figure}

\begin{figure}[t]
    \centering
    \includegraphics[viewport=0 0 737 238, width=8cm]{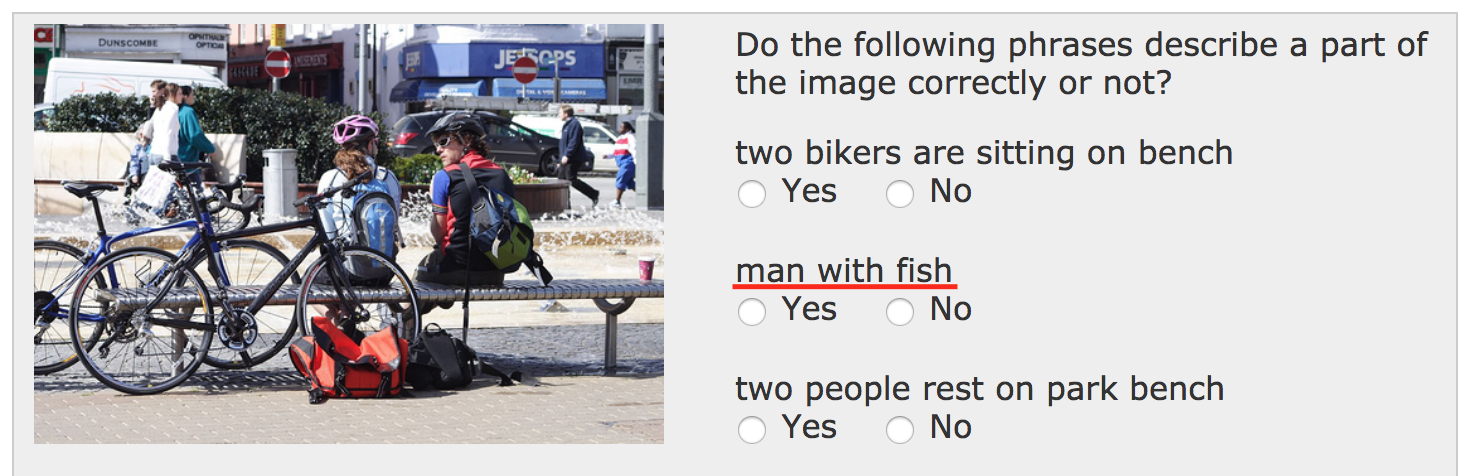}
    \vspace{-2mm}
    \caption{Example annotation tasks about an image in our interface. Note that the dummy question is highlighted with the red underline for readers' convenience; the interface does not have the underline.}
    \label{fig:interface}
\end{figure}

Visual relationships that could not be extracted via dependency parsing are manually annotated via a crowdsourcing service, Amazon Mechanical Turk (AMT). The screenshots of the instruction part and some annotation tasks in AMT are shown in Figures \ref{fig:instruction} and \ref{fig:interface}, respectively.
Our interface covers a single session for multiple annotation tasks. The first part of our interface provides the instructions for annotation, some annotation examples with correct and incorrect annotation results, and their reasons. It also notifies that there are dummy questions and workers who fail to answer these dummy questions would be rejected. Our dummy questions are the same annotation task but are preliminarily annotated so that we can automatically evaluate the dummy questions.  

Below the instruction part, the interface continues to actual annotation tasks. It shows 10 panels (on average) like Figure \ref{fig:interface}, each of which covers a single image (and thus have multiple annotation tasks for the image). Each image yields 1 to 10 candidate visual relationships to be annotated. We put many panels as a single session, which contains 50 annotation tasks. An annotation task is simply choosing \textit{Yes}/\textit{No} on a radio button. A single session comes with 5 dummy questions. A worker who achieves an accuracy higher than 0.8 on the dummy questions is accepted and otherwise are rejected. Figure \ref{fig:interface} shows an example, where the second question about the visual relationship ``man with a fish'' is a dummy question. The reward for a single session was set to \$0.2, and we only recruited workers who have the AMT master qualification.

Each session serves as a human intelligence task (HIT) in AMT. The overall flow of our AMT-based annotation is as follows.
\begin{enumerate}
\setlength{\parskip}{0cm} 
\setlength{\itemsep}{0cm} 
    \item Issue HITs in AMT.
    \item Workers annotate HITs.
    \item The accuracy of a single HIT is computed as the accuracy over the dummy questions.
    \item If the accuracy is higher than 0.8, we accept the HIT. Otherwise we reject it.\footnote{Only workers whose HIT is accepted get the reward. Rejected HITs yield no reward and are not included in the dataset.}
    \item The responses from the workers are aggregated and only candidate visual relationships for which the majority (\ie, more than 3 out of 5) of the responses are \textit{Yes} are labelled as true; otherwise, they are labelled as false.
\end{enumerate}

We carefully annotated 149 candidate visual relationships (from the ones to be annotated by AMT) by ourselves to evaluate the dataset.
The accuracy, precision, recall, and f-score is summarized in Table \ref{tab:AMT_result}. We can see that the AMT workers yield a high F-score of 94.9\%. The third and fourth rows in Table \ref{tab:statistic} show the statistics of true and false visual relationships obtained by AMT. We can see that the number of true relationships obtained by AMT is comparable to that number obtained by dependency parsing, indicating the limitation of dependency parsing in visual relationship detection. In addition, nearly 70k false relationships are annotated by AMT. These visual relationships annotated by AMT can be used to give feedback to and thus improve dependency parsing. However, we leave it as future work.
\begin{table}[t]
    \centering
    \begin{tabular}{r r r r}\hline
 Accuracy & Precision & Recall & F-score  \\ \hline
90.2 & 96.3 & 93.5 & 94.9 \\ \hline
    \end{tabular}
    \vspace{-2mm}
    \caption{Evaluation of AMT workers' annotations (\%).}
    \vspace{+2mm}
    \label{tab:AMT_result}
\end{table}


\section{Annotation Error Analysis}


\begin{figure}[t]
\begin{minipage}{0.46\hsize}
  \begin{center}
    \includegraphics[viewport=0 0 501 376, height=2.5cm]{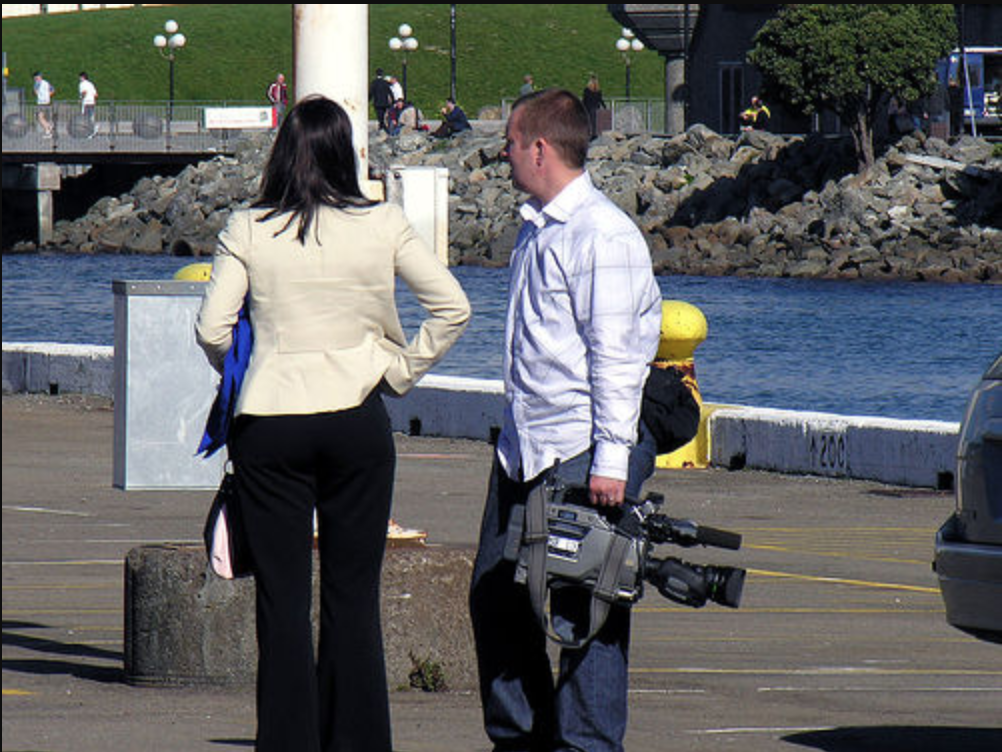}
    \\ (a) ``video camera standing next to a woman''
  \end{center}
  \end{minipage}
  \begin{minipage}{0.06\hsize}
        \hspace{2mm}
  \end{minipage}
 \begin{minipage}{0.46\hsize}
   \begin{center}
     \includegraphics[viewport=0 0 500 333, height=2.5cm]{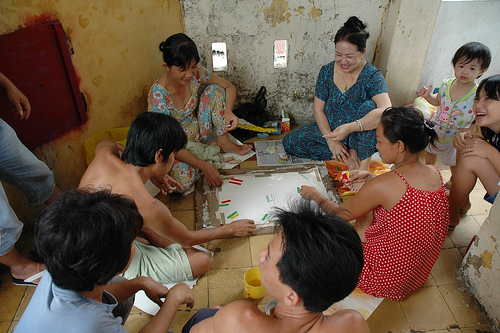}
     \\ (b) ``group of people playing board games''
    \end{center}
 \end{minipage}
 \begin{minipage}{0.46\hsize}
  \begin{center}
    \includegraphics[viewport=0 0 500 334, height=2.5cm]{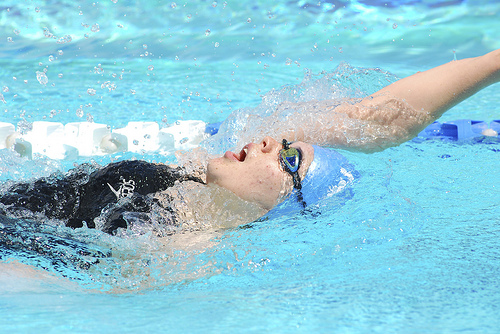}
    (c) ``backstroke in swimming pool''
  \end{center}
  \end{minipage}
 \begin{minipage}{0.06\hsize}
        \hspace{2mm}
 \end{minipage}
  \begin{minipage}{0.46\hsize}
  \begin{center}
    \includegraphics[viewport=0 0 500 333, height=2.5cm]{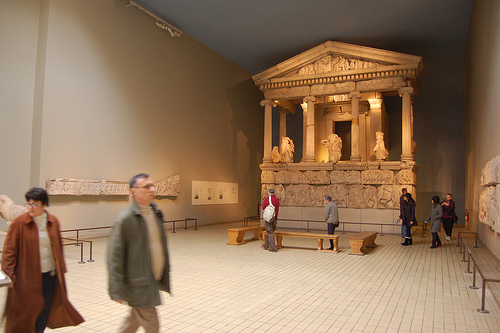}
    \\ (d) ``museum from antiquity''
  \end{center}
  \end{minipage}
  \caption{\label{fig:example} Examples of images and visual relationships with failed annotation.}
\end{figure}
Figures \ref{fig:example} (a) and (b) show two examples of dependency parsing failure to extract true visual relationships. Our dependency parsing extracted ``video camera standing next to a woman'' as a true visual relationship for Figure \ref{fig:example} (a), which actually is a false visual relationship. Similarly, For the image in Figure \ref{fig:example} (b), ``group of people playing board games'' is a true visual relationship but could not be extracted through dependency parsing. 
False positive visual relationships of dependency parsing could not be corrected because they were not included in candidate visual relationships to be annotated with AMT. We leave this issue as one of our future work.

Figures \ref{fig:example} (c) and (d) are failure examples in AMT-based annotation. ``backstroke in swimming pool'' is one of the candidate visual relationships extracted for Figure \ref{fig:example} (c), but ``backstroke'' is not an object and hence not a true visual relationship. However, all workers labeled it as true even though our instruction states that a coreference cannot be an entity, as shown in Figure \ref{fig:instruction}. This implies that our instruction is not sufficiently clear. For Figure \ref{fig:example} (d), ``museum from antiquity'' should be false because it shows an exhibition from antiquity but not a museum of antiquity. Yet, around 80\% of the workers labeled the candidate visual relationship as true. The workers might deduce that the exhibition should be in the museum and thus the candidate is true. However, the image itself does not show the museum itself, and so the museum cannot be an entity. 

\section{Conclusion}
In this paper, we constructed a dataset on visual relationships between two objects in images. The dataset was constructed by performing dependency parsing and AMT-based annotation on the Flickr30k entities dataset. Being different from previous studies, our dataset is annotated with both true and false visual relationships, covering all visual relationships in Flickr30k's captions. Our future work includes to explore a visual relationship detection model on our dataset.

\section*{Acknowledgement}
This work was supported by JST ACT-I.

\section*{References}
\bibliographystyle{lrec}
\bibliography{lrec2020W-xample}


\end{document}